# AcTED: Automatic Acquisition of Typical Event Duration for Semi-supervised Temporal Commonsense QA

**Felix Virgo**[1]   **Fei Cheng**[1]   **Lis Kanashiro Pereira**[2]
**Masayuki Asahara**[3]   **Ichiro Kobayashi**[4]   **Sadao Kurohashi**[1]

[1]Kyoto University
[2]NARA Institute of Science and Technology
[3]National Institute for Japanese Language and Linguistics
[4]Ochanomizu University
felix,feicheng,kuro@nlp.ist.i.kyoto-u.ac.jp, kanashiro.lis@is.naist.jp
masayu-a@ninjal.ac.jp, koba@is.ocha.ac.jp

## Abstract

We propose a voting-driven semi-supervised approach to automatically acquire the typical duration of an event and use it as pseudo-labeled data. The human evaluation demonstrates that our pseudo labels exhibit surprisingly high accuracy and balanced coverage. In the temporal commonsense QA task, experimental results show that using only pseudo examples of 400 events, we achieve performance comparable to the existing BERT-based weakly supervised approaches that require a significant amount of training examples. When compared to the RoBERTa baselines, our best approach establishes state-of-the-art performance with a 7% improvement in Exact Match.

**Keywords:** temporal commonsense QA, semi-supervised, voting-driven

## 1. Introduction

Understanding temporal commonsense such as how long an event typically lasts is essential in various NLP tasks like event timeline construction, narrative understanding question answering (Nakhimovsky, 1987; Cheng and Miyao, 2017; Ning et al., 2018; Leeuwenberg and Moens, 2019; Vashishtha et al., 2020; Cheng and Miyao, 2018; Cheng et al., 2020). However, we observe that when people mention the typical duration of an event, they tend to omit the time clues because it is considered commonsense knowledge. Figure 1 shows an example from MC-TACO (Zhou et al., 2019), a temporal commonsense QA task. In this example, the answer is not explicitly mentioned anywhere in the context, but we can infer using our commonsense that typically in a film, *music plays* for *minutes*, and not *years*. Consequently, existing weakly supervised approaches (Zhou et al., 2020; Yang et al., 2020; Virgo et al., 2022) struggle to learn the typical duration from the context. The extracted supervision leans towards explicit temporal signal (e.g. *for* in "strike *for* two days") and the substantial amount of the data hampers the learning efficiency.

Instead, our approach is motivated by the observation that the frequency distribution of an event over the whole duration units [1] in a corpus will be concentrated around those typical units. Therefore, by sampling sufficient sentences containing this event and utilizing a draft model to predict each duration label, we can eventually acquire an accurate estimation of the typical duration through majority voting. Even if the draft model makes a part of incorrect predictions, due to the majority being the typical duration, the final voting result would still be correct. We show an example at the bottom of Figure 1, which samples sufficient sentences containing the event *play music* from Wikipedia and acquires the typical *minutes* and *hours* units. Then we can leverage the typical duration label and its corresponding sentences as augmented pseudo data for improving the model performance in a semi-supervised fashion.

---
[1] In this paper, we define the set of duration units as {*seconds, minutes, hours, days, weeks, months, years, decades*}.

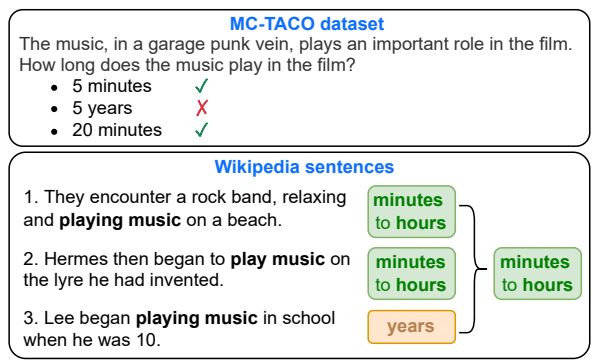

Figure 1: The top part is an example from MC-TACO asking the duration of the event *play music*. The bottom shows *play music* appears in Wikipedia sentences with the duration labels predicted by a draft model.

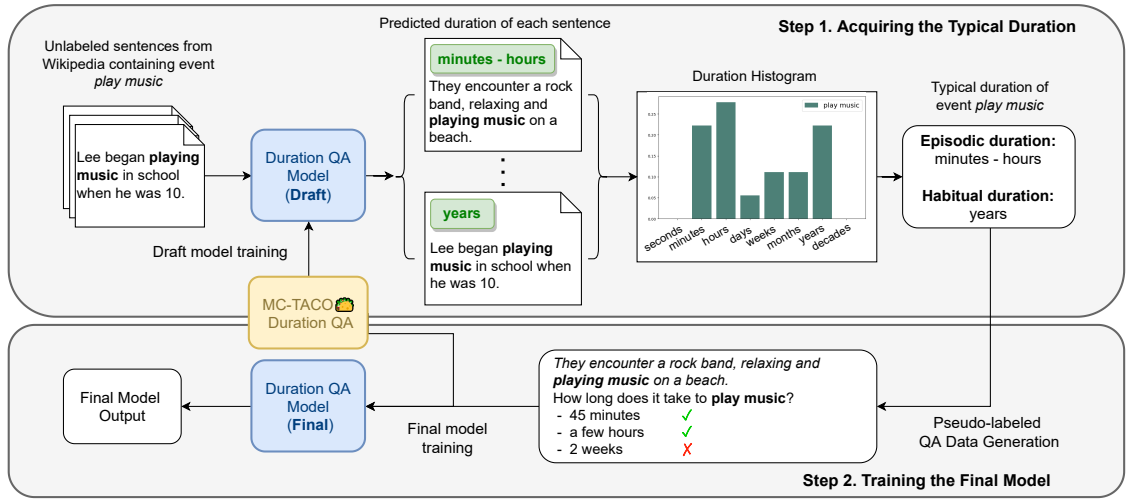

Figure 2: The semi-supervised approach for acquiring the typical duration of the event *play music* and generating the pseudo data for improving the temporal QA.

Mathew and Katz (2009) observed that the typical duration of an event can be distinguished into two categories, *episodic* refers to the typical period of an event occurring occasionally, while *habitual* means an event repeatedly occurs over a period of time. Williams and Katz (2012) manually annotated the binary habituality labels for classifying whether an event in a certain context is *episodic* or *habitual*. Surprisingly, our voting histogram exhibits a similar bi-modal characteristic, with many events showing two non-adjacent peaks across the duration units. For instance, *take a course* has one peak (*episodic*) in *hours* and another (*habitual*) in *months*. What sets our research apart from the previous studies is that our approach does not rely on costly human effort, and we easily extend the target events to the informative verb phrases, whereas two previous research focus on verb lemmas only.

In summary, we propose a novel semi-supervised approach, which leverages a voting strategy across the duration units to acquire typical duration labels. These pseudo labels exhibit surprisingly high accuracy and broad coverage in human evaluation. We then append the augmented pseudo data into training. In the MC-TACO task, our models, trained with pseudo examples of only hundreds of events, show superior performance to state-of-the-art weakly supervised models with significantly less amount of data. [2]

## 2. Related Work

**Duration question answering and duration knowledge acquisition.** Zhou et al. (2019) create MC-TACO, a temporal commonsense QA dataset, consisting of questions from five temporal phenomena, including *typical duration* [3]. Pan et al. (2011) manually annotated the TimeBank corpus (Pustejovsky et al., 2003) with their expected durations by specifying upper and lower bounds. Gusev et al. (2011); Williams and Katz (2012) extracted explicit event duration at the lemma level from the web using unsupervised patterns. Mathew and Katz (2009); Williams and Katz (2012); Friedrich and Pinkal (2015) proposed a supervised classification task for distinguishing *episodic* or *habitual* event with human annotated labels. Cheng and Miyao (2020) annotated the date anchor of events with the beginning and ending information, by which the duration can be automatically induced.

**Semi-supervised learning.** Semi-supervised learning often trains a draft model on supervised data for predicting unlabeled data and selecting high-confidence predictions as pseudo-data for augmentation. He et al. (2020) study noisy pseudo data for improving the machine translation and summarization tasks. Du et al. (2021) uses task-specific query embeddings from labeled data to retrieve pseudo-labeled web sentences. Our approach is distinguished from these studies without relying on confidence scores.

## 3. Method

Figure 2 shows the overall process of our proposed method. Our method automatically acquires the typical duration of various events by aggregating the duration predictions of various sentences from the web. The duration prediction is done using a draft model, a pre-trained language model fine-tuned on MC-TACO-duration data. The acquired typical duration is then used for generating pseudo-labeled data, which is used to train the final model.

---

[2] We will release the code and the dataset to the public upon the acceptance of the paper.

[3] We following the previous work to conduct the experiemnts on the duration part, named MC-TACO-duration.

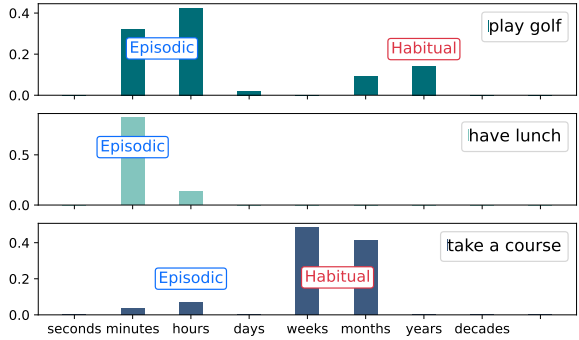

Figure 3: Automatically acquired duration histograms of three events.

**Duration QA model.** To train the draft and final models, we fine-tune a pre-trained language model, such as BERT (Devlin et al., 2019) or RoBERTa (Liu et al., 2019). The model receives two elements: (1) the context sentence concatenated with the question and (2) a candidate answer, separated with a special token (`[SEP]` token in BERT). The final hidden state of the first token in the sequence (`[CLS]` token in BERT) is fed into a dense output layer to make a binary prediction on each instance, *plausible* or *not plausible*.

**Step 1. Acquiring Typical Duration of Events through Majority-Voting.** We first collect phrases from the ConceptNet [4] entries. Then we select those parsed [5] as verb phrases to be events.

For each event, we sample $50$ sentences from Wikipedia that contain it. We group the events that share the same verb lemma and noun, e.g., *playing music* and *played music*, as one event *play music*. To predict the duration of an event with various context sentences using the draft model, we formulate it as a QA task similar to MC-TACO. Given a context sentence, a question formulated as: "*How long does it take to [event] ?*," and a candidate answer, the task is to predict whether the answer is plausible or not. We use all $8$ duration units as the answers, i.e., *seconds*, *minutes*, *hours*, *days*, *weeks*, *months*, *years*, and *decades*.

We aggregate the predicted duration of each event across the $50$ sentences from Wikipedia to construct a duration histogram for voting. To determine the typical duration, we take the peak duration unit from the duration distribution of the histogram. A peak is defined when the number of predictions in a duration unit is larger than its two neighboring units. A duration distribution usually has one or two peaks in it. Moreover, we also take the peak's neighboring unit as part of the typical duration if the number of its predictions is $\geq 75\%$ of the peak's predictions.

If the histogram is bi-modal, we naturally interpret two peaks as *episodic duration* (smaller units) and *habitual duration* (larger units). In the case of the single-peak histogram, we consider it as *episodic duration*. Figure 2 shows an example of two peaks in the aggregated predictions of *play music*, which are on *minute-hour* level and *year* level. In the sentence: "*They encounter a rock band, relaxing and playing music on a beach*," the event *play music* is an occasionally occurring event that typically takes minutes or hours (*episodic*). Meanwhile, in the sentence: "*Lee began playing music in school when he was 10*," the event *play music* can take years since it repeatedly occurs over a period of time (*habitual*). Figure 3 shows the duration histogram examples of three events automatically acquired by our method. For example, *have lunch* has a single-peak distribution at the *minutes* (*episodic*), while in *play golf*, we can observe two peaks at the *hours* (*episodic*) and *years* (*habitual*).

**Step 2. Training the Final Model** First, we generate Pseudo-labeled data for QA. The pseudo-labeled data follows the same format as MC-TACO. For each acquired typical duration label, we randomly select one corresponding sentence from Step 1. For each context, we formulate the question as: *How long does it take to* `[event]` *?* For each question, we generate $3$ positive answers and $4$ negative answers, formulated as: `[number] [duration unit]`. The positive answers use the acquired *episodic duration* as their `[duration unit]`. For the negative answers, we randomly select the `[duration unit]` where it is at least two units apart from the positive answers. Suppose that the positive answer is in *minutes* then the negative answers cannot be in *seconds* or *hours*. We choose two units apart since the adjacent temporal units are also likely to be the temporal units of an event (Pan et al., 2011). For both answers, the `[number]` is generated randomly between the range of each unit, e.g., from $1$ to $23$ for *hours*. We also randomly perturb a part of answers by replacing `[number]` with a quantifier, e.g., "a few".

## 4. Experiments and Discussion

**Experiment Settings** MC-TACO-duration consists of the $126$-question train set and the $314$-question test set. For the final model, we first fine-tune BERT on the pseudo-labeled data with $n$ event questions where $n \in \{100, 200, 400, 1000\}$. We then fine-tune the model on MC-TACO-duration. Different $n$ in the pseudo-labeled data is trained with a similar number of steps ($\sim 400$ steps). We then fine-tune the model on MC-TACO-duration for 250 steps. We use a batch size of 32 and a learning rate of $1e-5$ with cross-entropy loss and Adam (Kingma and Ba, 2015) as the optimizer.

---
[4] https://conceptnet.io/
[5] https://github.com/allenai/allennlp

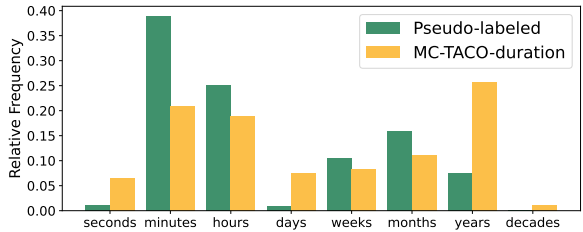

Figure 4: Duration distribution of positive answers in the pseudo-labeled data and the MC-TACO-duration.

| Model | EM | F1 |
|---|---|---|
| *BERT$_{base}$ based* | | |
| BERT$_{base}$ | 32.27 | 58.74 |
| TACOLM (Zhou et al., 2020) | 34.60 | - |
| E-Pred (Yang et al., 2020) | **39.68** | 63.63 |
| AcTED$_{base}$ ($n = 100$) | 35.99 | 61.20 |
| AcTED$_{base}$ ($n = 200$) | 36.31 | 63.39 |
| AcTED$_{base}$ ($n = 400$) | 37.90 | **64.10** |
| AcTED$_{base}$ ($n = 1000$) | 35.67 | 63.71 |
| *RoBERTa$_{large}$ based* | | |
| RoBERTa$_{large}$ | 40.45 | 67.42 |
| UDS-T (Virgo et al., 2022) | 45.86 | 70.52 |
| AcTED$_{large}$ ($n = 100$) | 44.16 | 68.45 |
| AcTED$_{large}$ ($n = 200$) | 47.03 | 71.80 |
| AcTED$_{large}$ ($n = 400$) | 48.73 | 70.34 |
| AcTED$_{large}$ ($n = 1000$) | **49.15** | **72.30** |

Table 1: Performances on MC-TACO-duration. $n$ is the number of events in the pseudo-labeled data. All the scores are the 3-run average of different random seeds.

| Model | EM | F1 |
|---|---|---|
| *Confidence-based method* | | |
| BERT$_{base}$ | 33.65 | 61.44 |
| RoBERTa$_{large}$ | 45.65 | 69.98 |
| *Vote-based method (ours)* | | |
| BERT$_{base}$ | **35.67** | **63.71** |
| RoBERTa$_{large}$ | **49.15** | **72.30** |

Table 2: Comparison to Confidence-based Semi-supervised approach with the same 1,000 event pseudo data.

We use two evaluation metrics to measure the model performance: (1) Exact Match (EM), which measures how many questions a system can correctly label all candidate answers, and (2) F1, which measures the average overlap between predictions and the ground truth.

**Examining event habituality distribution of MC–TACO.** It is important to investigate the real distribution of the event habituality in MC-TACO-duration. By sampling 100 questions (23%) from the whole MC-TACO-duration and manually counting the number of *episodic* and *habitual* events, we found that over 90% of the central events in the questions of MC-TACO are *episodic*.

**Investigating the quality of pseudo data.** Figure 4 shows the duration distribution of our pseudo-data (1,000 events) and MC-TACO-duration (126 + 314 questions). MC-TACO, which is artificially curated, has more events that lasted *years* compared to our pseudo-data which has more short-duration events, i.e., *minutes* and *hours*. The pseudo labels exhibit a broad coverage of duration units and appear to resemble the real-world distribution more. For instance, we often mention the events evolved in those short units of *minutes* and *hours* more.

We also conducted an intrinsic evaluation for our typical duration data using crowdworkers from Amazon Mechanical Turk. We randomly sampled 147 events of our typical duration data and asked the annotators if the duration of a given event (without being given any context sentence) makes sense or not. Each event was evaluated by 3 annotators and labeled using the majority vote. Our duration data achieves 93% accuracy for the *episodic* du-ration and 77% accuracy for the *habitual* duration, showing that our method can accurately capture the typical duration of events. Considering that *episodic* has higher accuracy and *habitual* is more likely accompanied by explicit temporal clues, we generate the pseudo data only from *episodic* peak.

**Main Results.** Table 1 shows the results on the MC-TACO-duration of our proposed models and several mentioned weakly supervised models. Our proposed models improve the Exact Match (EM) score up to 8.7 points (22%) and the F1 score up to 4.9 points (7%) compared to the baseline RoBERTa-large model. Our best model outperforms the highest published result, i.e., RoBERTa-large-based UDS-T, with 7% EM improvement. The performances generally increase as the number of events increases, which suggests the potential effectiveness of our method at scale. We plan to experiment with more sentences and more events in future work.

In terms of the comparison of efficiency, our models achieve comparable or better results with only hundreds of event questions for training. TACOLM and E-Pred retrieve training instances with rule-based patterns relying on specific knowledge of the target language, while UDS-T recast the human-annotated data. All three methods require more training events than ours, i.e., $1.5$M for TACOLM, $31$K for E-Pred, and $7$K for UDS-T.

We further compare our voting-driven approach to the conventional confidence-based semi-supervised approach. Table 2 shows that our method shows superior performance in both the BERT$_{base}$ and RoBERTa$_{large}$ settings.

## 5. Conclusion

We propose a novel semi-supervised method for automatically voting the typical duration labels of phrase-level events for solving MC-TACO-Duration. Our methods achieve state-of-the-art RoBERTa-based results with only hundreds of events. Experimental results further suggest the potential effectiveness of our method at scale.

## Limitations

A bias towards a certain style of sentences might occur since our method relies on sampling sentences from one web source, i.e., Wikipedia. Using a larger number of sentences from various sources might alleviate this issue.

Yang et al. (2020) show the importance of the numerical value of duration. In the pseudo-labeled QA data generation, we currently choose random `[number]` in the answers, which does not portray the accurate numerical information of the duration. This can be improved by considering the duration distribution itself. For example, if *hours* is the peak and the number of predictions in *minutes* is higher than *days*, it might suggest that it is more plausible for the event to last $1$ or $2$ hours rather than $22$ or $23$ hours.

## 6. Bibliographical References

Fei Cheng, Masayuki Asahara, Ichiro Kobayashi, and Sadao Kurohashi. 2020. Dynamically updating event representations for temporal relation classification with multi-category learning. In *Findings of the Association for Computational Linguistics: EMNLP 2020*, pages 1352–1357, Online. Association for Computational Linguistics.

Fei Cheng and Yusuke Miyao. 2017. Classifying temporal relations by bidirectional LSTM over dependency paths. In *Proceedings of the 55th Annual Meeting of the Association for Computational Linguistics (Volume 2: Short Papers)*, pages 1–6, Vancouver, Canada. Association for Computational Linguistics.

Fei Cheng and Yusuke Miyao. 2018. Inducing temporal relations from time anchor annotation. In *Proceedings of the 2018 Conference of the North American Chapter of the Association for Computational Linguistics: Human Language Technologies, Volume 1 (Long Papers)*, pages 1833–1843, New Orleans, Louisiana. Association for Computational Linguistics.

Fei Cheng and Yusuke Miyao. 2020. Predicting event time by classifying sub-level temporal relations induced from a unified representation of time anchors.

Jacob Devlin, Ming-Wei Chang, Kenton Lee, and Kristina Toutanova. 2019. BERT: Pre-training of deep bidirectional transformers for language understanding. In *Proceedings of the 2019 Conference of the North American Chapter of the Association for Computational Linguistics: Human Language Technologies, Volume 1 (Long and Short Papers)*, pages 4171–4186, Minneapolis, Minnesota. Association for Computational Linguistics.

Jingfei Du, Edouard Grave, Beliz Gunel, Vishrav Chaudhary, Onur Celebi, Michael Auli, Veselin Stoyanov, and Alexis Conneau. 2021. Self-training improves pre-training for natural language understanding. In *Proceedings of the 2021 Conference of the North American Chapter of the Association for Computational Linguistics: Human Language Technologies*, pages 5408–5418, Online. Association for Computational Linguistics.

Annemarie Friedrich and Manfred Pinkal. 2015. Automatic recognition of habituals: a three-way classification of clausal aspect. In *Proceedings of the 2015 Conference on Empirical Methods in Natural Language Processing*, pages 2471–2481, Lisbon, Portugal. Association for Computational Linguistics.

Andrey Gusev, Nathanael Chambers, Divye Raj Khilnani, Pranav Khaitan, Steven Bethard, and Dan Jurafsky. 2011. Using query patterns to learn the duration of events. In *Proceedings of the Ninth International Conference on Computational Semantics (IWCS 2011)*.

Junxian He, Jiatao Gu, Jiajun Shen, and Marc'Aurelio Ranzato. 2020. Revisiting self-training for neural sequence generation. In *International Conference on Learning Representations*.

Diederik P. Kingma and Jimmy Ba. 2015. Adam: A method for stochastic optimization. In *3rd International Conference on Learning Representations, ICLR 2015, San Diego, CA, USA, May 7-9, 2015, Conference Track Proceedings*.

Artuur Leeuwenberg and Marie-Francine Moens. 2019. A survey on temporal reasoning for temporal information extraction from text. *Journal of Artificial Intelligence Research*, 66:341–380.

Artuur Leeuwenberg and Marie-Francine Moens. 2020. A survey on temporal reasoning for temporal information extraction from text (extended abstract). *CoRR*, abs/2005.06527.

Yinhan Liu, Myle Ott, Naman Goyal, Jingfei Du, Mandar Joshi, Danqi Chen, Omer Levy, Mike Lewis, Luke Zettlemoyer, and Veselin Stoyanov. 2019. RoBERTa: A Robustly Optimized BERT Pretraining Approach. *arXiv e-prints*, page arXiv:1907.11692.

Thomas Mathew and Graham Katz. 2009. Supervised categorization of habitual and episodic sentences. In *Sixth Midwest Computational Linguistics Colloquium*, Bloomington, Indiana.

Alexander Nakhimovsky. 1987. Temporal reasoning in natural language understanding: The temporal structure of the narrative. In *Proceedings of the Third Conference on European Chapter of the Association for Computational Linguistics*, EACL '87, page 262–269, USA. Association for Computational Linguistics.

Qiang Ning, Hao Wu, and Dan Roth. 2018. A multi-axis annotation scheme for event temporal relations. In *Proceedings of the 56th Annual Meeting of the Association for Computational Linguistics (Volume 1: Long Papers)*, pages 1318–1328, Melbourne, Australia. Association for Computational Linguistics.

Feng Pan, Rutu Mulkar-Mehta, and Jerry R. Hobbs. 2011. Annotating and learning event durations in text. *Computational Linguistics*, 37(4):727–752.

Jason Phang, Thibault Févry, and Samuel R. Bowman. 2018. Sentence encoders on stilts: Supplementary training on intermediate labeled-data tasks. *CoRR*, abs/1811.01088.

James Pustejovsky, Patrick Hanks, Roser Saurí, Andrew See, Rob Gaizauskas, Andrea Setzer, Dragomir Radev, Beth Sundheim, David Day, Lisa Ferro, and Marcia Lazo. 2003. The timebank corpus. *Proceedings of Corpus Linguistics*.

Peng Shi and Jimmy Lin. 2019. Simple bert models for relation extraction and semantic role labeling. *ArXiv*, abs/1904.05255.

Siddharth Vashishtha, Adam Poliak, Yash Kumar Lal, Benjamin Van Durme, and Aaron Steven White. 2020. Temporal reasoning in natural language inference. In *Findings of the Association for Computational Linguistics: EMNLP 2020*, pages 4070–4078, Online. Association for Computational Linguistics.

Felix Virgo, Fei Cheng, and Sadao Kurohashi. 2022. Improving event duration question answering by leveraging existing temporal information extraction data. In *Proceedings of the Thirteenth Language Resources and Evaluation Conference*, pages 4451–4457, Marseille, France. European Language Resources Association.

Jennifer Williams and Graham Katz. 2012. Extracting and modeling durations for habits and events from twitter. In *ACL*.

Zonglin Yang, Xinya Du, Alexander Rush, and Claire Cardie. 2020. Improving event duration prediction via time-aware pre-training. In *Findings of the Association for Computational Linguistics: EMNLP 2020*, pages 3370–3378, Online. Association for Computational Linguistics.

Ben Zhou, Qiang Ning, Daniel Khashabi, and Dan Roth. 2020. Temporal common sense acquisition with minimal supervision. In *ACL*.

## 7. Language Resource References

Zhou, Ben and Khashabi, Daniel and Ning, Qiang and Roth, Dan. 2019. *"Going on a vacation" takes longer than "Going for a walk": A Study of Temporal Commonsense Understanding*. Association for Computational Linguistics.